\documentclass{article}

\usepackage[utf8]{inputenc}
\usepackage{PRIMEarxiv}
\usepackage{hyperref}
\usepackage{url}
\usepackage{amsfonts}
\usepackage{graphicx}
\usepackage{booktabs}
\usepackage{fancyhdr}
\usepackage{lipsum}
\usepackage{tikz}
\usetikzlibrary{shapes,arrows.meta,positioning}
\usepackage{algorithm}
\usepackage{algpseudocode}
\usepackage{appendix}
\usepackage{titletoc}

\begin{document}

\title{Crowdsourced Adaptive Surveys}
\author{Yamil Ricardo Velez \\ Department of Political Science, Columbia University \\ Email: \url{yrv2004@columbia.edu}}
\date{\today}

\pagestyle{fancy}
\fancyhead[LO]{Crowdsourced Adaptive Surveys}
\fancyhead[RE]{Yamil Ricardo Velez}

\maketitle

\begin{abstract}
Public opinion surveys are vital for informing democratic decision-making, but responding to rapidly evolving information environments and measuring beliefs within niche communities can be challenging for traditional survey methods. This paper introduces a crowdsourced adaptive survey methodology (CSAS) that unites advances in natural language processing and adaptive algorithms to generate question banks that evolve with user input. The CSAS method converts open-ended text provided by participants into survey items and applies a multi-armed bandit algorithm to determine which questions should be prioritized in the survey. The method's adaptive nature allows for the exploration of new survey questions, while imposing minimal costs in survey length. Applications in the domains of Latino information environments, national issue importance, and local politics showcase CSAS's ability to identify topics that might otherwise escape the notice of survey researchers. I conclude by highlighting CSAS's potential to bridge conceptual gaps between researchers and participants in survey research.
\end{abstract}

\newpage

\section{Introduction}

Survey research plays a critical role in informing political decision-making \cite{page1983effects}. However, surveys can be too slow to adapt to changing information environments, especially in societies marked by cultural and political heterogeneity. First, surveys may struggle to identify and prioritize emerging issues. This is particularly evident in the context of elections, where campaign-related events, foreign policy incidents, or unexpected economic conditions are typical occurrences. The inherent lag time between recognizing an emerging concern and fielding a survey can lead to missed opportunities for capturing real-time opinion shifts. Second, there is often uncertainty about which items from a larger universe of items should be included in a survey. Third, participants may not fully grasp the measures used in traditional surveys. This issue is especially acute when studying minority groups, whose perspectives and experiences may diverge from mainstream viewpoints.\footnote{For instance, \cite{anoll2018makes} shows that minority groups’ lived experiences and collective beliefs can lead to different understandings of concepts like political engagement.}

In this paper, I propose a crowdsourced adaptive survey methodology (CSAS) that leverages advances in natural language processing and adaptive algorithms to create participant-generated questionnaires that evolve over time. I use open-ended responses from participants to create question banks comprised of potential survey questions, from which questions are prioritized using a multi-armed bandit algorithm. While the survey is in the field, participants contribute to the question bank and respond to existing questions, enabling the algorithm to explore and prioritize survey questions that resonate with the study population. Even in well-trodden settings such as identifying important issues, the CSAS method produces promising items that warrant further investigation (see Figure 1 for a summary of the process and representative issues recovered using the method).

I apply this methodology across three domains: gauging the prevalence of misinformation within minority communities, evaluating issue importance in the aggregate, and identifying local political concerns. First, I use CSAS to develop an evolving battery of issues to gauge issue importance in the aggregate. Despite seeding the algorithm with Gallup issue categories, I find that popular issue topics based on open-ended responses depart from the set of ``most important issues,'' reflecting concerns over healthcare, inflation, political accountability, and crime.\footnote{While some Gallup codes (e.g., "ethics, moral, and family decline") are not expressed in everyday language, complicating the comparison with CSAS-generated summaries, many codes (e.g., "immigration," "poverty," "crime") closely align with CSAS outputs.} Then, I move to more niche applications: misinformation within the Latino community and local political concerns. I find that CSAS reveals claims and issues that would likely escape the notice of survey researchers. 

The advantages of CSAS are threefold. First, it enables survey researchers to capture trends in public opinion in real-time, reflecting the public's evolving beliefs and concerns. Second, it allows survey researchers to apply a more inductive approach to questionnaire construction. Finally, it democratizes the survey process by allowing respondents to contribute to instruments. These benefits come at little cost in terms of survey length. Researchers can set the number of ``dynamic questions'' in advance, and select the appropriate algorithm for determining how questions should be prioritized. For example, one can rely on a set of tried-and-true items, while allocating a few survey slots for dynamic questions.

\begin{figure}[h]
\centering
\label{fig: fig_intro}
\begin{tikzpicture}[node distance=2cm and 1cm, auto]

\tikzstyle{process} = [rectangle, minimum width=3cm, minimum height=1cm, text centered, text width=7cm, draw=black, fill=orange!30]
\tikzstyle{arrow} = [thick,->,>=stealth]
\tikzstyle{column} = [rectangle, rounded corners, minimum width=2.5cm, minimum height=8cm, text centered, draw=black, text width=3.5cm, fill=blue!30]

\node (in1) [process] {Gather Open-ended Responses};
\node (pro1) [process, below of=in1] {Convert Responses to Structured Questions};
\node (pro2) [process, below of=pro1] {Apply Toxicity and Redundancy Filters};
\node (pro3) [process, below of=pro2] {Participants Respond to Own Question and K Other Questions};
\node (pro4) [process, below of=pro3] {Update Question Bank using Algorithm};

\draw [arrow] (in1) -- (pro1);
\draw [arrow] (pro1) -- (pro2);
\draw [arrow] (pro2) -- (pro3);
\draw [arrow] (pro3) -- (pro4);

\node (col1) [column, right=1cm of pro2] {
    \begin{tabular}{l}
    Inflation \\ \\
    Immigration \\ \\
    Economy \\ \\
    Unifying the \\ country \\ \\
    Race \\ Relations \\ \\
    Poverty/Hunger/\\Homelessness \\ \\
    Crime/Violence \\ \\
    Moral/Ethical/\\Family decline \\ \\
    \end{tabular}
};
\node (col2) [column, right=.5cm of col1, fill=green!30] {
    \begin{tabular}{l}
    Cost of Living \\ \\
    Healthcare \\ Affordability \\ \\
    Economic \\ Stability \\ \\
    Political \\ Accountability \\ \\
    Education \\ System \\ \\
    Freedom of \\ Speech \\ \\
    Democracy \\ \\
    Abortion
    \end{tabular}
};

\node[above=0.5cm of in1, anchor=center] {\textbf{CSAS Process}};
\node[above=0.5cm of col1, anchor=center] {\textbf{Gallup Items}};
\node[above=0.5cm of col2, anchor=center] {\textbf{CSAS Items}};

\end{tikzpicture}
\caption{CSAS Process Flowchart and Representative Set of Issues uncovered by CSAS.}
\end{figure}
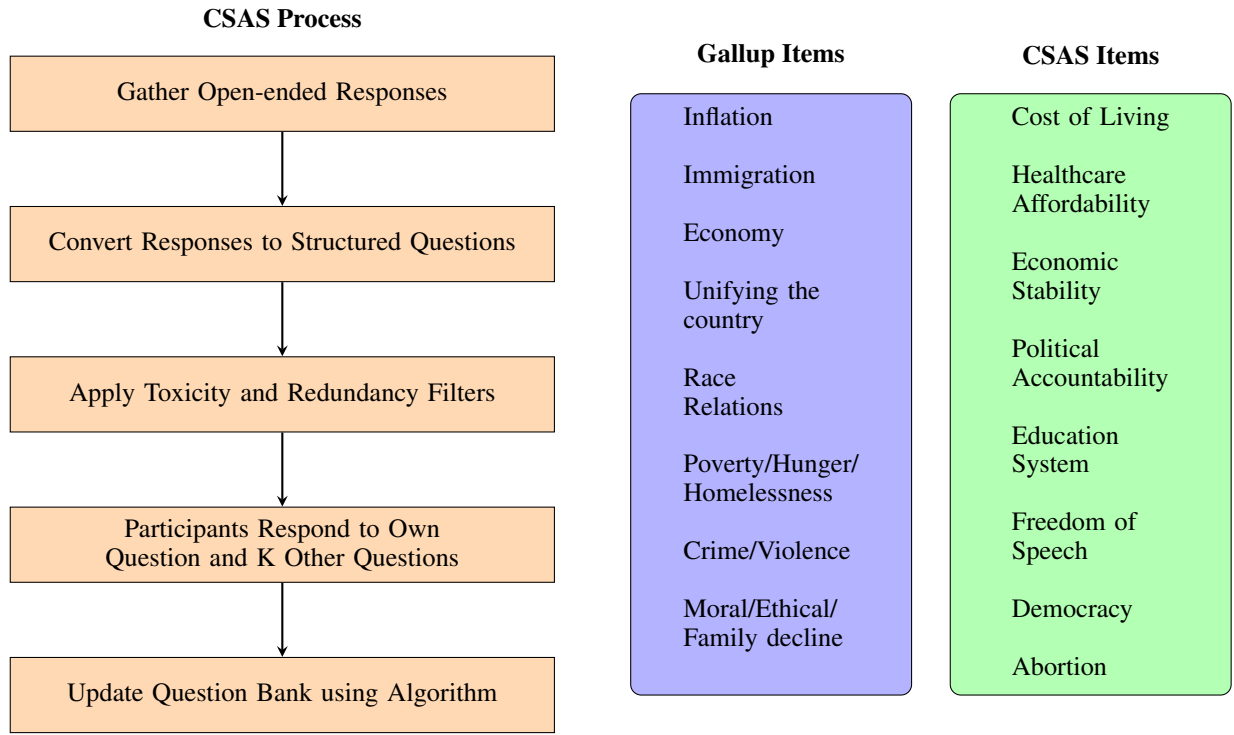

Beyond applications to misinformation and issue importance, the method can be applied to research topics such as social movements, public reactions to unfolding political events, political representation, or the concerns of minority groups. By enabling researchers to adapt survey instruments to changing information environments and democratizing the survey process, the CSAS approach has the potential to provide novel insights and complement traditional survey methods across various sub-fields in political science.

\section{Dynamic Survey Methodologies: Existing Approaches}

Influential texts on survey design stress the importance of a ``tailored'' approach to recruitment and stimuli, where survey materials are adapted to populations under study \cite{dillman2014internet}. Adapting questionnaires to respondents can enhance measurement and satisfaction. For instance, branching questions can reduce respondent burden and measurement error by eliminating irrelevant sub-questions \cite{krosnick1993comparisons, dillman2005survey}. Questions measuring recognition of elected officials and voting in sub-federal elections can be tailored using location to produce more relevant question content via ``piped in'' text that is automatically shown to participants in online surveys \cite{ansolabehere2009guide}. These examples illustrate how surveys already possess dynamic elements that respond to user input or data.

\subsection{Computerized Adaptive Tests}

Scholars have recently developed methods for carrying out computerized adaptive tests (CATs) in public opinion surveys \cite{montgomery2013computerized, montgomery2020so, montgomery2022adaptive}. CAT algorithms ``respond to individuals’ prior answers by choosing subsequent questions that will place them on the latent dimension with maximum precision and a minimum number of questions.'' (p. 173). CATs are typically employed using latent scales, where the goal is to optimize the number of questions. Montgomery and Cutler show that CATs offer a superior approach to traditional static batteries, and these tools can be easily implemented in survey software such as Qualtrics \cite{montgomery2022adaptive}.

CATs rely on pre-existing measurement scales (e.g., political knowledge, personality batteries). However, in settings where the objective is to capture novel issues or changes in the information environment, scholars and practitioners may want to learn about the prevalence of discrete beliefs, some of which cannot be known in advance. Thus, while CATs allow us to enhance precision when estimating latent traits, there are settings where the question bank cannot be fixed in advance and describing the nature of discrete items, rather than estimating positions on latent scales, is the primary objective. These settings include misinformation research, where survey questions draw from fact-checking sources and social media data, and studies of elections and campaigns, where events and shifts in news coverage are integral to understanding the dynamics of races. 

\subsection{Wiki Surveys}

Wiki surveys are a collaborative survey methodology where users help shape instruments \cite{salganik2015wiki}. Drawing inspiration from online information aggregation portals such as Wikipedia, wiki surveys balance three principles: greediness, collaborativeness, and adaptivity. Greediness refers to capturing as much information as respondents are willing to provide, collaborativeness refers to allowing respondents to modify instruments (e.g., proposing new items), and adaptivity refers to optimizing instruments to select the ``most useful information.'' While wiki surveys have shown promise in facilitating collective decision making (e.g., allowing users to vote on policies -- both pre-determined and user-generated -- that should be considered by local governments), existing applications rely on pairwise comparisons between options provided by survey designers and participants. However, pairwise comparisons may not be useful in settings where options can be accorded the same weight, the decision is not zero-sum, and outcomes can be more accurately measured on an ordinal or continuous scale.

\section{The Crowdsourced Adaptive Survey Method}

Building on the wiki survey and other attempts to insert dynamic elements into existing surveys (e.g., CAT approaches), I develop a crowdsourced adaptive survey (CSAS) method that enables question banks to evolve based on user input and does not impose constraints on question formats. I use generative pre-trained transformers (GPTs) to convert participants' open-ended responses into structured questionnaire items (see \cite{velez2024confronting} for an example). I then employ adaptive algorithms \cite{offer2021adaptive} to identify the 'best-performing' questions from the question bank.\footnote{Questions are deemed 'best-performing' based on their ability to maximize researcher-defined objectives (e.g., mean accuracy estimates, mean importance ratings).} First, each respondent answers an open-ended question about a given topic that is cleaned, summarized, and converted into a structured survey question format. Second, respondents respond to their own questions, along with \emph{k} other questions from a question bank generated by previous participants. Finally, ratings for user-submitted questions and \emph{k} questions drawn from the question bank are updated using a multi-armed bandit algorithm, adjusting the probabilities of presenting these questions to future participants in subsequent surveys.\footnote{Following the ``greediness'' principle of wiki surveys, users rate their own questions to maximize the information collected about items.}

\begin{table}[h]
\centering
\caption{Overview of the CSAS Methodology}
\label{tab:csas_overview}
\renewcommand{\arraystretch}{1.5} 
\begin{tabular}{|p{3cm}|p{3cm}|p{5cm}|}
\hline
\textbf{Step} & \textbf{Model} & \textbf{Details} \\ \hline
Collect Open-ended Responses & - & Participants provide responses to open-ended questions. \\
Process Open-Ended Responses & LLM (e.g., OpenAI's GPT-3.5) & Open-ended responses are processed and converted into structured survey questions using LLMs. \\
Apply Filters & Document Embeddings (e.g., OpenAI's \emph{text-embedding-ada-002}) & \textbf{If flagged as redundant,} do not add to question bank. \\
& Toxicity Detection (e.g., OpenAI's moderation endpoint) & \textbf{If flagged as toxic,} do not add to question bank. \\
& & \textbf{If not flagged as redundant or toxic,} add structured text to question bank. \\
Participant Ratings & - & Participants rate their own question and \emph{k} other questions selected using a multi-armed bandit. \\
Update Question Bank & Multi-Armed Bandit Algorithm & The question bank is updated using ratings, prioritizing the most highly rated questions. \\ \hline
\end{tabular}
\end{table}

The three essential features of the proposed method are open-ended questions, a question bank, and an algorithm for updating question selection probabilities. Table \ref{tab:csas_overview} displays the different steps of the CSAS method: eliciting potential items using open-ended questions; processing and filtering candidate items; and optimizing question selection. I walk through each step in turn. 

The open-ended question is used to query participants in a free-form manner about a given topic, issue, or claim. Since these data will be unstructured, introducing heterogeneity on dimensions such as length, style, and grammar, a response conversion stage is typically necessary. For example, given that Likert scale questions in public opinion surveys tend to be brief, one can convert an open-ended response of arbitrary length into a sentence-long summary using a GPT. GPTs are large-parameter language models that can perform various tasks such as text prediction, summarization, and classification at levels that mirror human performance \cite{vaswani2017attention, radford2018improving}. Recent years have seen the development of a diverse range of GPTs, with proprietary models like OpenAI's GPT-4, Anthropic's Claude, and Google's Gemini demonstrating robust capabilities across various text manipulation, summarization, and generation tasks. In parallel, ``open source'' and ``open weights'' models, including Meta's Llama and Mistral AI's Mixtral, have emerged, offering competitive performance in similar domains (see Appendix E for a detailed discussion and direct comparison of these models).

Once open-ended response data are in a usable, structured format, they can be included in a question bank. Though inclusion of questions can be unrestricted at this stage, researchers may want to impose additional constraints to reduce redundant questions and apply filters to ensure that survey questions meet the researcher's objectives (e.g., reducing toxicity, increasing relevance). 

Focusing on redundancy first, if two respondents submit responses about Democratic spending priorities with only minor differences in wording, it may be unnecessary to include both questions. Moreover, multi-armed bandit algorithms can struggle to identify the best-performing arm when arms are equally matched (i.e., a ``no clear winner'' scenario). Assuming near-identical questions are rated similarly, this increases the odds of failing to identify the best-performing item \cite{offer2021adaptive}. Given that open-ended responses are unlikely to be \emph{exact} duplicates, filtering using more sophisticated NLP methods such as document embeddings can be useful. Document embeddings locate texts on a high-dimensional space and can be used to identify similarities between texts \cite{rheault2020word}. By applying document embeddings, researchers can quantify the similarity between different questions even when the wording varies, and retain only questions that surpass a pre-defined threshold of similarity. 

Researchers may also want to apply additional filters to ensure that questions meet pre-specified criteria on dimensions such as relevance and toxicity. For example, for a survey of issue importance, a survey researcher may choose to exclude responses referring to the personal characteristics of politicians. Given that this is a classification task, one may opt for a supervised learning model trained on a labeled dataset or a GPT model, among other options. The same holds for identifying and removing toxic responses. Given that a small percentage of respondents resort to ``trolling'' or toxic behavior, ensuring other participants are not exposed to harmful content is paramount \cite{lopez2018why}. 

The next challenge lies in how questions are presented and selected within the survey, balancing between identifying ``best-performing items'' (exploitation) and examining the full set of candidate items (exploration). This entails the use of an algorithm that takes a set of candidate items and determines how those items are presented to participants. In some domains, such as misinformation research or issue salience studies, prioritizing popular claims or important issues may be crucial. Conversely, in fields like personality research, capturing the full spectrum of trait variation, including low-prevalence items, is more important. In the latter case, uniform sampling from the question bank could be useful.

Across the three applications presented in this paper, I focus on identifying the ``best-performing'' survey questions. I employ multi-armed bandit algorithms, specifically Gaussian Thompson sampling, to systematically evaluate the quality of questions over time. These algorithms balance two key objectives: exploitation (focusing on questions that show promise) and exploration (testing additional questions to assess their potential). The algorithms use outcome measures to guide the assignment of participants to different questions, prioritizing those that score higher on predetermined metrics. Through this iterative process, Thompson sampling efficiently allocates more participants to questions that ``resonate'' more with the sample.

\section{An Application to Issue Importance}

In contrast to conventional approaches that use pre-defined issues to study issue importance  (see \cite{ryan2023issue} for a critique), CSAS can be used to produce a dynamic slate of issues. This can be helpful in estimating support for idiosyncratic issues that may not appear on the national agenda, but still inspire strong reactions among ``issue publics'' \cite{elliott2020democracy} or serve as issues that could be mobilized in future elections, corresponding to the elusive notion of ``latent opinion'' described in \cite{key1961public}.

Since 1935, Gallup's Most Important Problem (MIP) question has been used to identify the issue priorities of the American public. Using an open-ended format, participants are asked ``What do you think is the most important problem facing this country today?,'' with responses being hand-classified into a set of categories corresponding to broad issue or policy areas. Despite its adoption in public opinion research, the measure has been criticized for being an imperfect proxy of salience. As \cite{wlezien2005salience} argues, the question asks respondents to provide information on two distinct concepts — importance and ``problem status.'' While some respondents may interpret the question as one where they can offer a personally relevant issue, others may interpret it as an opportunity to highlight a problem affecting the nation as a whole. 

More recently, \cite{ryan2023issue} make a case for moving beyond fixed closed-ended questions and hand-coded classifications based on open-ended questions when measuring issue importance. Like the MIP, \cite{ryan2023issue} use open-ended questions to elicit issue positions from participants. However, they ask participants to directly reflect on issues of personal importance. Moreover, in contrast to the typical MIP method, they obtain closed-ended measures of personal importance for the issues elicited using the open-ended method. This approach provides a richer amount of information about the \emph{degree} to which an elicited issue may hold importance to an individual. Recent studies in this vein have recovered high levels of stability \cite{velez2024confronting} and sizable causal effects for this ``core issue'' in conjoint settings \cite{ryan2023issue, velez_2023_tradeoffs_in_latino_politics}.

Applying CSAS in this context is straightforward. Participants report personally relevant issues using open-ended responses\footnote{The wording of the open-ended question is based on the question described in \cite{ryan2023issue}.}; open-ended responses are transformed into a structured question and included in a question bank; and participants respond to the different issue topics in an issue importance battery, with Gaussian Thompson sampling used to optimize question selection. I seed the question bank with a set of eight Gallup items that were popular at the time and assess whether the crowdsourced issue topics receive higher importance ratings. The issues were the following: 'Immigration', 'Economy', 'Race Relations', 'Poverty', 'Crime', 'Ethics, Moral, and Family Decline', 'Unifying the country', and 'Inflation.'

From September 11 until September 13, 2023, I collected data from a national quota sample balanced on age, race, and gender using CloudResearch Connect (N=820). CloudResearch Connect provides a low-cost, high-quality online convenience sample. Its demographic representativeness is comparable to other popular sample providers such as Prolific \cite{stagnaro2024representativeness}.\footnote{CloudResearch Connect participants score at the upper end of attentiveness, which may not reflect all survey environments. For samples with lower attentiveness, researchers could implement additional measures such as conditioning inclusion of participant-generated questions on attention check performance or using open-ended response quality checks.}  OpenAI's GPT-4 was instructed to convert unstructured text into issue topics and filter out irrelevant or redundant topics.\footnote{When rating their own issues, the issue ``room temperature superconductors'' was presented to those submitting gibberish or nonsense responses.} This was accomplished using a retrieval-augmented generation pipeline. This process involved retrieving the five nearest neighbors for each potential new topic, allowing GPT-4 to consider similarity and avoid redundancy when generating new issues. For each submitted question, I used OpenAI's \emph{ada} embeddings model to generate a 1536-dimensional embeddings vector for each item and retrieved the five nearest neighbors using a vector database.\footnote{I used Pinecone's API, a vector database service, to store and retrieve embeddings.} GPT-4 was instructed to avoid generating issues similar to the ``nearest neighbors'' that were retrieved (see Appendix D). 

Gaussian Thompson Sampling (GTS) was then used to determine which questions to present to subsequent participants. Though traditional Thompson Sampling requires binary outcomes due to its use of the Beta distribution \cite{offer2021adaptive}, GTS can be leveraged to accommodate continuous outcomes \cite{agrawal2017near}.\footnote{As in other research \cite{offer2021adaptive}, a probability floor of .01 was employed to guarantee that every item in the item bank retained a non-zero chance of being presented to participants.} GTS was implemented in real-time using a custom-built back-end system created by the author. In contrast to previous applications of adaptive experiments in political science that have leveraged batch-wise Thompson sampling, probabilities were adjusted at the respondent level. 

Recent studies have shown that the efficacy of GPTs such as GPT-4 in generating issue categories using open-ended text is on par with classification algorithms trained on thousands of examples, achieving performance levels marginally below that of human evaluators \cite{mellon2024ais}. Each participant rated their own issue, along with eight others that were determined using GTS. Issue importance was measured using the question, "How important is this issue to you personally?" Responses were recorded on a five-point ordinal scale with the following options: "Not at all important," "Slightly important," "Moderately important," "Very important," and "Extremely important."

\subsection{Results}

\begin{figure}
    \centering
    \includegraphics[scale=.65]{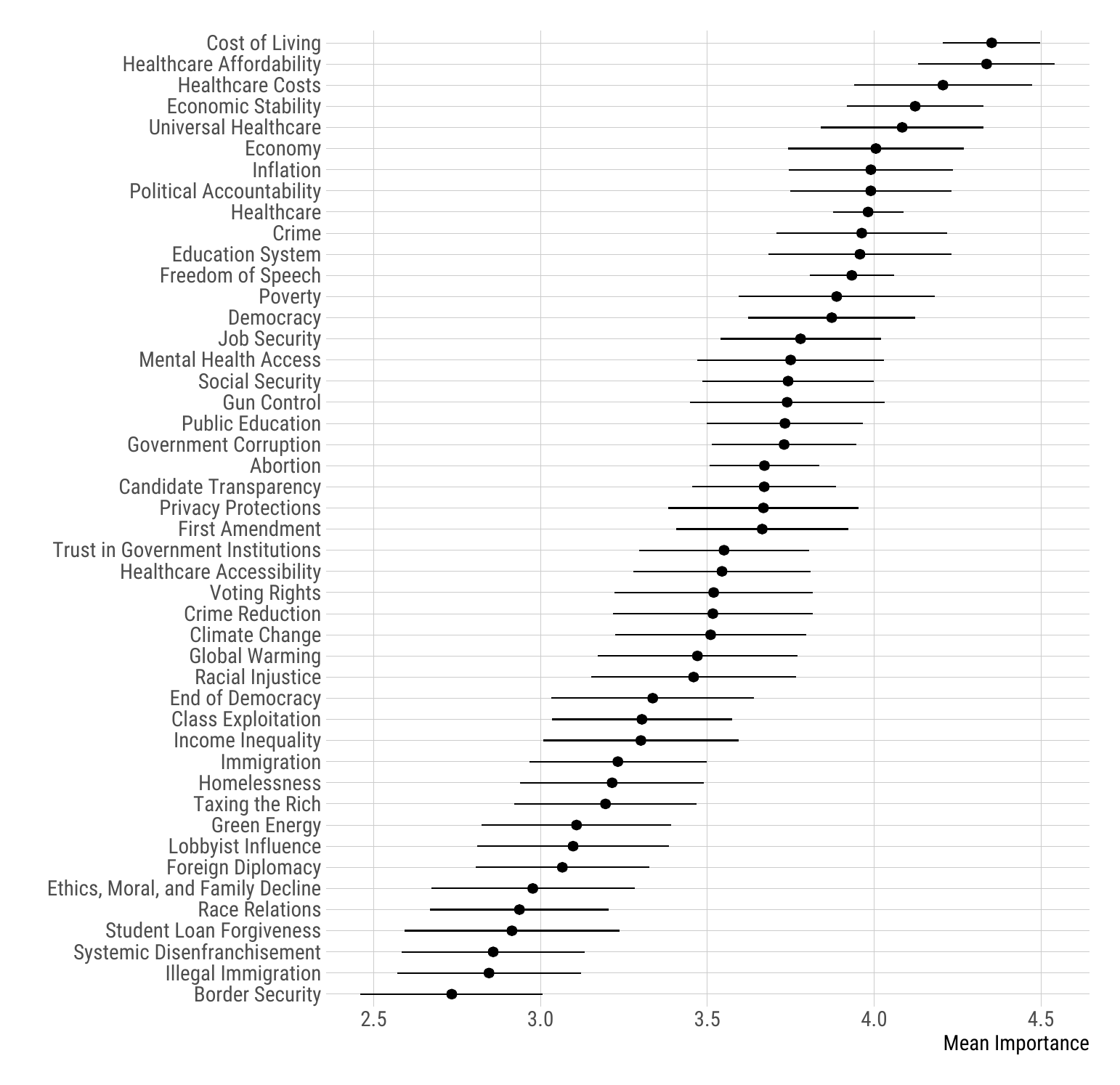}
    \caption{IPW-weighted estimates of survey questions measuring issue importance with corresponding 95\% confidence intervals.}
    \label{fig: fig3}
\end{figure}

Figure \ref{fig: fig3} displays mean estimates for issue topics receiving 50 or more issue importance ratings.\footnote{Mean estimates are weighted by the inverse probability of selection, as in \cite{offer2021adaptive}.} As shown in the figure, the highest-rated issues were focused on the economy and health care, with issues such as ``Cost of Living,'' ``Healthcare Affordability,'' ``Healthcare Costs,'' ``Economic Stability,'' and ``Universal Healthcare.'' Issues rated lower on importance include more social and culture issues such as immigration and voting rights.\footnote{Though one might argue that the healthcare-relevant issues should be collapsed into a single item, allowing for more refined issue categories allows researchers to explore different dimensions of the same issue that may be important to participants. Moreover, if more coarse issue categories are desirable, these can be incorporated into the LLM's prompt using few-shot prompting (e.g., including multiple examples in the prompt that produce the intended classifications).} The MIP issue topics of ``Race Relations'' and ``Ethics, Moral, and Family Decline'' appeared among the lower-rated items, as did topics related to immigration (i.e., ``Illegal Immigration,'' ``Border Security''). In the list of highly rated topics, we see issues that would likely not appear in traditional issue importance batteries such as ``Mental Health Access,'' ``Privacy Protections,'' and ``Candidate Transparency.'' Moreover, the frequent mention of various economic and healthcare dimensions is instructive, the prominence of socioeconomic considerations in the sample. In Appendix B, I assess heterogeneity across partisan subgroups finding some issues that generate consensus across parties such as healthcare costs, political polarization, and economic issues. 

\section{An Application to Latino Information Environments}

Moving from issue importance to concerns within niche communities, I use the CSAS method to identify rumors, negative political claims, and misinformation reaching Latinos, a group that has received attention among journalists and social scientists due to potential misinformation campaigns targeting the community \cite{cortina2022conspiratorial, velez2023latino}. I focus on Latinos for two reasons. First, fact-checking is still a relatively new institution within Latino-oriented media \cite{velez2023latino}. Existing organizations might overlook important claims that circulate within the community due to resource constraints and the possibility that best practices for verification are still being developed. Second, private, encrypted messaging applications used by Latinos such as WhatsApp and Telegram may hinder the detection of false claims. In contrast to misinformation that is transmitted through social media such as Instagram, Twitter, and Facebook, fact-checkers and researchers may not be privy to claims circulating in private channels. 

Implementing this more ``bottom-up'' approach to misinformation detection,  I fielded a survey of 319 self-identified Latinos from the United States using the survey platform, CloudResearch Connect, from July 6-7, 2023. First, participants were asked two open-ended question regarding negative claims they had heard about Republicans and Democrats.\footnote{This approach expands the inclusion criteria for the question bank beyond the traditional definition of misinformation to include falsifiable statements that portray parties or candidates negatively. While unconventional, this broader inclusion criteria allows us to cast a wider net, revealing interesting distinctions between newsworthy scandals and false claims that might otherwise go unnoticed. Future iterations could employ more sophisticated methods, such as comparing open-ended responses against databases of fact checks to isolate verifiably false statements. Additionally, one could explore alternative objectives, like identifying items that are factually true but commonly misperceived as false.} These claims were then passed to a fine-tuned version of OpenAI's \emph{ada} text generation model that classified the text as a ``verifiable claim.''\footnote{Note that there is a distinction between the ada embeddings and text generation models.} Fine-tuning was necessary to ensure that questions entering the question bank were falsifiable political claims, rather than value judgments (e.g., politicians are evil). To carry out the fine-tuning step, a mixture of researcher-provided examples and participant-provided examples (N=87) were hand-coded to indicate whether a claim was falsifiable in principle. Hand-coded classifications were then used to fine-tune the \emph{ada} text generation model.\footnote{Since this study was conducted, more advanced models have emerged. These models may obviate the need for fine-tuning through the use of few-shot prompting \cite{chen2023prompting}, where a selection of examples can be included in the prompt by the researcher to guide the model.}

I also used a similarity and toxicity filter before adding items to the question bank.\footnote{If a user submission did not successfully pass these filters, respondents did not rate their own submission, but instead were asked to rate generic items about Republicans (Democrats) being too conservative (liberal).} For each submitted question, I used the \emph{ada} model in OpenAI's embeddings API, retrieved the five nearest neighbors using a vector database, and filtered out questions with a similarity score above .90.\footnote{In initial tests before data collection, lower similarity thresholds such as .80 were found to produce false positives (e.g., classifying ``Biden is a tax cheat'' and ``Trump is a tax cheat'' as sufficiently similar) and higher similarity thresholds such as .95 produced false negatives (see Appendix H for a discussion of different methods for reducing redundant items).} I also used OpenAI's moderation endpoint to filter out offensive and toxic claims. Claims that passed these filtering steps were added to the question bank.

After the submission and cleaning step, participants responded to their own question bank submission, along with four items from the question bank and six items capturing conspiracy beliefs and more common misinformation items (e.g., Covid-19 vaccines modify your DNA). All of the questions were presented in a matrix format, with a four-point accuracy scale. The response options were ``not at all accurate,'' ``not very accurate,'' ``somewhat accurate,'' and ``very accurate.'' Four claims were taken from the front pages of Latino-oriented fact-checking websites (i.e., Telemundo's T-Verifica, Univision's El Detector) to seed the question bank. As in the issue importance study, GTS was used to update question selection probabilities with accuracy ratings being used as the metric for updates.

\section{Results}

\begin{figure}
    \centering
    \includegraphics[scale=.29]{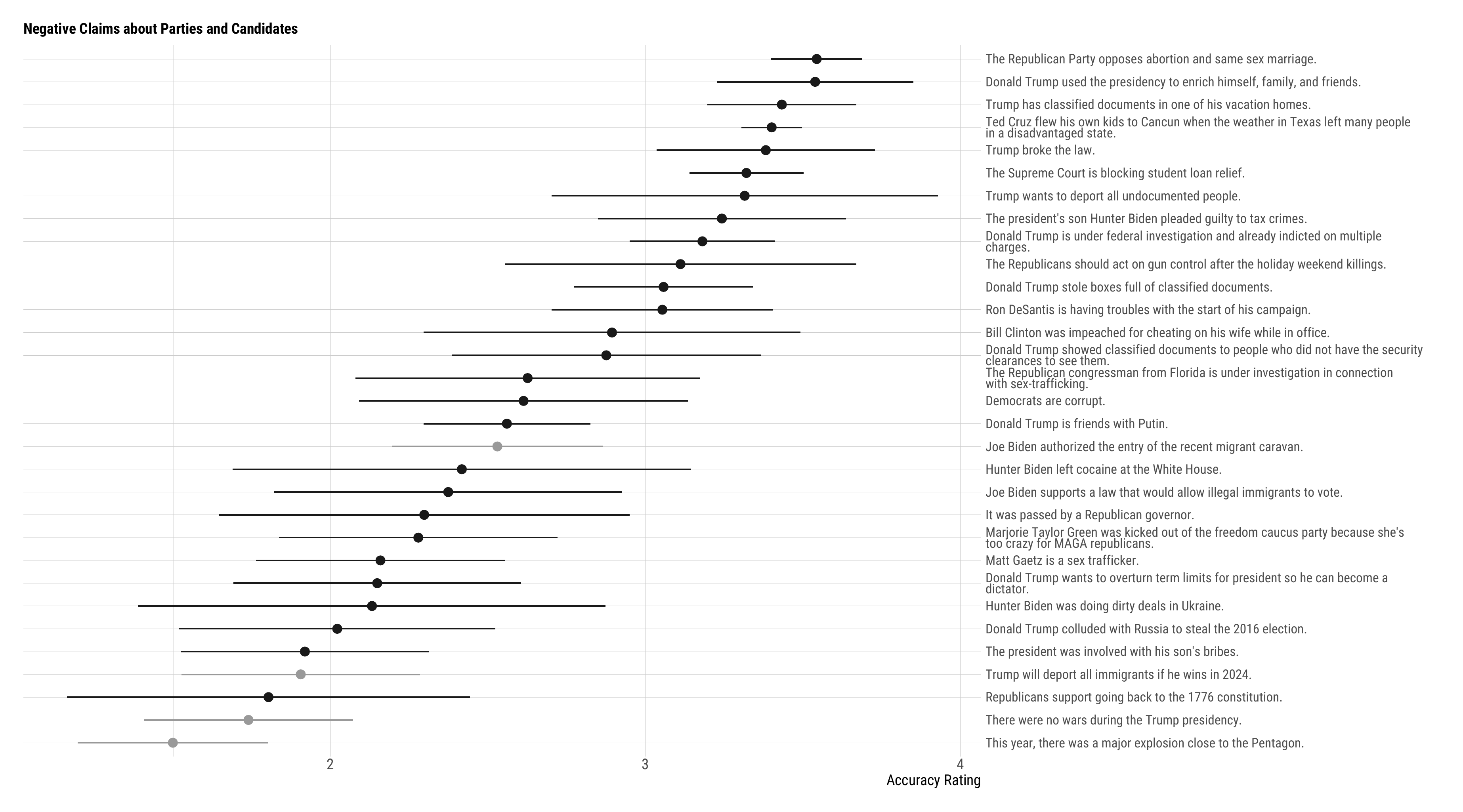}
    \caption{IPW-weighted estimates of survey questions measuring negative political claims with corresponding 95\% confidence intervals. Items in gray were initial seed items based on fact-checked claims produced by Latino-focused fact-checking organizations. Black items are participant-generated items.}
    \label{fig: fig1}
\end{figure}

Figure \ref{fig: fig1} displays mean accuracy estimates for claims receiving more than ten ratings. The claims with the highest perceived factual accuracy covered information about both Republicans and Democrats. Party stereotypes about Republican positions on moral issues were rated as highly accurate ($\bar{x}$ = 3.54; SE = .07), as were claims that Trump used the presidency to enrich family and friends ($\bar{x}$ = 3.54; SE = .159) and possessed classified documents in his vacation homes ($\bar{x}$ = 3.43; SE = .121). Other highly rated claims focused on Republicans such as Ted Cruz ($\bar{x}$ = 3.40; SE = .049), extreme policy positions such as a claim that Trump plans to deport all undocumented people ($\bar{x}$ = 3.32; SE = .31), and President Biden's son, Hunter Biden ($\bar{x}$ = 3.24; SE = .201). The lowest-rated claims typically involved false statements, allegations, or extreme descriptions of issue positions. Claims that scored especially low on perceived accuracy included ``This year, there was a major explosion close to the White House'' ($\bar{x} = 1.50$; SE = .15), ``There were no wars during the Trump presidency'' ($\bar{x} = 1.74$; SE = .17), ``Republicans support going back to the 1776 constitution.'' ($\bar{x} = 1.80$; SE = .33), ``Trump will deport \textbf{all} immigrants if he wins in 2024'' ($\bar{x} = 1.91$; SE = .19), and ``The president was involved with his son's bribes'' ($\bar{x} = 1.92$; SE = .20). 

Whereas claims rated highly on perceived accuracy mostly reflected actual events or generalizations of party positions, claims with lower accuracy ratings typically involved verifiably false information or oversimplifications. These findings are instructive in that they reveal a level of discernment in the aggregate. Objectively false claims are generally seen as less credible by participants. Instead, higher accuracy ratings are observed when participants judge claims that are widely reported in the news (e.g., Ted Cruz's cancun trip) or that reflect commonly-held perceptions of party positions (e.g., Republicans opposing abortion and same sex marriage). Though the initial seed claims based on fact-checks were small in number, the analysis revealed a surprising disparity: the most readily believed claims were not identified by the fact-checking organizations, but rather originated from other participants. In Appendix B, I explore heterogeneity by levels of trust in the private, encrypted messaging application WhatsApp, which has been argued to be a vector of misinformation \cite{velez2023latino}. I find that those who trust the platform are less likely to view genuine newsworthy scandals as accurate, but rate factually inaccurate claims similarly to those who do not trust WhatsApp. This suggests that gaps between the two groups might not be a function of misinformation \emph{per se}, but instead more general political knowledge.

The results underscore the complexity of the information environment among Latinos in the United States. They engage with a variety of narratives, some of which portray different parties negatively, but also reflect actual events. This study provides a valuable snapshot of how a prominent and politically pivotal ethnic group, which receives significant attention from campaigns, interfaces with information about political parties. However, it is important to note that these findings may not generalize directly to other contexts. An advantage of CSAS is that these group-specific understandings may be reflected in the question bank and researchers can assess whether questions that perform well within certain subgroups also perform well in others. Moreover, while the focal outcome of this study is perceived accuracy, one could implement variations of CSAS for misinformation measurement that optimize for exposure. In this setup, participants might be asked to report whether they have seen a set of claims. Though exposure does not equate to belief, this approach could help identify claims that may become more prevalent over time.

\section{An Application to Local Political Issues}

In Appendix K, I examine local political concerns. Similar to the case of studying minority opinion, inconsistent polling at the local and state levels can complicate the development of quality survey items. Dovetailing with recent research on the ``nationalization'' of local politics \cite{hopkins2018increasingly}, policy domains that are typically considered national in scope such as immigration, gun policy, foreign policy (e.g., stances on Gaza), and the environment emerge as user-submitted items and gain traction within the sample. In surveys of particular states, Congressional districts, or cities, CSAS could be a valuable tool for uncovering local attitudes, beliefs, and preferences. 

\section{Concerns and Caveats}

\subsection{Is the CSAS method compatible with traditional survey design?} 

Despite the limitations of traditional surveys in identifying changing information environments or measuring responses within niche communities, the two approaches are not at odds. Researchers can decide the number of adaptive questions, and include these questions in standard batteries. For example, in the Latino survey, participants rated a pre-existing set of false claims and conspiracies, along with an adaptive set, in a question matrix. Before using this method, scholars should determine whether the marginal benefit of having a designated slot for exploratory questions is worth the survey time and cost.\footnote{In the 2024 ANES pilot, the median response time for an open-ended issue importance question was 17 seconds.} A distinct advantage of multi-armed bandits is that several items can be explored despite having a smaller set of survey slots. The two approaches can also work in tandem when there are multiple phases of data collection. An initial wave (or pilot) could use CSAS to develop a fixed battery of questions for future waves, functioning much like pilot studies that gather open-ended data to inform scale construction \cite{weller1998structured}. With CSAS, however, future surveys can be designed not only with open-ended content in hand, but question ratings and posterior probabilities that a given question is the best-performing question. This approach may be optimal if researchers prefer to split their research process into exploratory and confirmatory stages, as is recommended in \cite{egami2018make}. Pre-registration across these different stages could lend more credibility to conclusions derived using this method (see \cite{offer2022battling} for an example). 

\subsection{Late Arrivals, Prompting Participants, and Costs of LLM Inference}

In Appendix C, I address additional issues such as ``late arrivals'' or items added to the question bank toward the end of the survey, general concerns about crafting open-ended questions, and inference costs. I estimate inference costs for both closed-source and open-source models, finding that researchers should expect to pay between 0.005 and 0.01 cents per participant when using this method. Furthermore, I discuss the issue of toxicity and recommend using moderation APIs. In Appendix G, I provide detailed steps for implementing the CSAS method. I develop a Django application that can be hosted on popular platforms such as Amazon Web Services, Replit, or Google Cloud. Finally, in Appendix I, I examine whether sample composition varies over the course of the study but fail to detect evidence of ``demographic bias.'' I discuss potential solutions that could be implemented in settings where this may be an issue. 

\section{Conclusion}

This paper introduces a novel adaptive survey methodology that engages participants in stimuli creation. Applied first to national issue importance, the CSAS method enabled participants to generate questions reflecting prioritized issues. While traditional priorities like economic conditions, healthcare, crime, and education emerged as top-rated issues, the method also surfaced unique concerns such as candidate transparency and privacy protections. That user-submitted questions frequently outranked those based on Gallup issue definitions suggests CSAS could identify public priorities that conventional survey approaches might miss.

The method was also used to study misinformation within the Latino community. The analysis revealed that the most highly rated claims were partisan stereotypes, accurate statements, or widely reported allegations. In contrast, claims that were rated lower on accuracy were often objectively false and reflected more blatant misinformation. This application highlights CSAS’s utility in identifying contested claims that might otherwise go unnoticed in traditional surveys. In another setting involving the measurement of more niche concerns, CSAS was used to examine local political concerns, where its flexibility enabled the identification of both locally salient issues absent from national discourse and national issues not traditionally associated with local government.

A distinct advantage of the adaptive design is that the exploration of new items can yield a larger number of items than one would measure in a traditional static survey. By dynamically adjusting the question bank, CSAS can explore new issues (e.g., artificial intelligence) and unexpected areas of agreement across partisan or ideological subgroups. Given heterogeneity in ratings across individuals, applying more sophisticated algorithms such as ``contextual bandits'' could improve estimates within subgroups \cite{offer2022battling}. Although no substantial over-time sample imbalances were detected in this study, future applications could employ deconfounded Thompson sampling to adjust for potential variations in demographic composition over time \cite{qin2022adaptive}.\footnote{In settings where measures exhibit temporal volatility, particularly in longitudinal settings, time-aware adaptive algorithms could account for temporal drift in estimates \cite{cavenaghi2021non}.}

Future research employing CSAS has the potential to explore a variety of topics where participant-generated content is particularly valuable. For instance, in candidate choice, voters not only consider issue positions or party affiliation, but also personal attributes such as honesty and competence \cite{lenz2012follow}. CSAS could be useful in uncovering additional elements that influence voting decisions. Furthermore, when identifying norms, core beliefs, or key sources of identity within hard-to-reach communities, CSAS could provide significant insights, reducing the misalignment of researcher-defined concepts with respondents' actual perceptions. 

CSAS could also be used to develop measurement scales that more accurately reflect considerations important to participants. Many constructs in the social sciences reflect latent variables (e.g., democracy, gentrification, identity), and the CSAS method could serve as a useful tool for extracting folk definitions of these various concepts. This kind of research, which prioritizes the perspectives of study populations, could not only bridge the gap between researchers and respondents, but could also produce significant advances in our understanding of public opinion and political behavior, more broadly.

\newpage

\bibliographystyle{unsrt}
\bibliography{references}

\begin{thebibliography}{10}

\bibitem{page1983effects}
Benjamin~I Page and Robert~Y Shapiro.
\newblock Effects of public opinion on policy.
\newblock {\em American political science review}, 77(1):175--190, 1983.

\bibitem{anoll2018makes}
Allison~P Anoll.
\newblock What makes a good neighbor? race, place, and norms of political participation.
\newblock {\em American Political Science Review}, 112(3):494--508, 2018.

\bibitem{dillman2014internet}
Don~A Dillman, Jolene~D Smyth, and Leah~Melani Christian.
\newblock {\em Internet, phone, mail, and mixed-mode surveys: The tailored design method}.
\newblock John Wiley \& Sons, 2014.

\bibitem{krosnick1993comparisons}
Jon~A Krosnick and Matthew~K Berent.
\newblock Comparisons of party identification and policy preferences: The impact of survey question format.
\newblock {\em American Journal of Political Science}, pages 941--964, 1993.

\bibitem{dillman2005survey}
Don~A Dillman and Leah~Melani Christian.
\newblock Survey mode as a source of instability in responses across surveys.
\newblock {\em Field methods}, 17(1):30--52, 2005.

\bibitem{ansolabehere2009guide}
Stephen Ansolabehere and Brian Schaffner.
\newblock Guide to the 2008 cooperative congressional election survey.
\newblock {\em Harvard University, draft of February}, 9, 2009.

\bibitem{montgomery2013computerized}
Jacob~M Montgomery and Josh Cutler.
\newblock Computerized adaptive testing for public opinion surveys.
\newblock {\em Political Analysis}, 21(2):172--192, 2013.

\bibitem{montgomery2020so}
Jacob~M Montgomery and Erin~L Rossiter.
\newblock So many questions, so little time: Integrating adaptive inventories into public opinion research.
\newblock {\em Journal of Survey Statistics and Methodology}, 8(4):667--690, 2020.

\bibitem{montgomery2022adaptive}
Jacob~M Montgomery and Erin~L Rossiter.
\newblock {\em Adaptive Inventories: A Practical Guide for Applied Researchers}.
\newblock Cambridge University Press, 2022.

\bibitem{salganik2015wiki}
Matthew~J Salganik and Karen~EC Levy.
\newblock Wiki surveys: Open and quantifiable social data collection.
\newblock {\em PloS one}, 10(5):e0123483, 2015.

\bibitem{velez2024confronting}
Yamil~Ricardo Velez and Patrick Liu.
\newblock Confronting core issues: A critical assessment of attitude polarization using tailored experiments.
\newblock {\em American Political Science Review}, pages 1--18, 2024.

\bibitem{offer2021adaptive}
Molly Offer-Westort, Alexander Coppock, and Donald~P Green.
\newblock Adaptive experimental design: Prospects and applications in political science.
\newblock {\em American Journal of Political Science}, 65(4):826--844, 2021.

\bibitem{vaswani2017attention}
Ashish Vaswani, Noam Shazeer, Niki Parmar, Jakob Uszkoreit, Llion Jones, Aidan~N Gomez, {\L}ukasz Kaiser, and Illia Polosukhin.
\newblock Attention is all you need.
\newblock {\em Advances in neural information processing systems}, 30, 2017.

\bibitem{radford2018improving}
Alec Radford, Karthik Narasimhan, Tim Salimans, Ilya Sutskever, et~al.
\newblock Improving language understanding by generative pre-training.
\newblock {\em OpenAI}, 2018.

\bibitem{rheault2020word}
Ludovic Rheault and Christopher Cochrane.
\newblock Word embeddings for the analysis of ideological placement in parliamentary corpora.
\newblock {\em Political Analysis}, 28(1):112--133, 2020.

\bibitem{lopez2018why}
Jesse Lopez and D.~Sunshine Hillygus.
\newblock Why so serious?: Survey trolls and misinformation.
\newblock {\em SSRN}, 3 2018.

\bibitem{ryan2023issue}
Timothy~J Ryan and J~Andrew Ehlinger.
\newblock Issue publics: How electoral constituencies hide in plain sight.
\newblock {\em Elements in Political Psychology}, 2023.

\bibitem{elliott2020democracy}
Kevin~J Elliott.
\newblock Democracy's pin factory: Issue specialization, the division of cognitive labor, and epistemic performance.
\newblock {\em American Journal of Political Science}, 64(2):385--397, 2020.

\bibitem{key1961public}
Vladimir~O Key.
\newblock Public opinion and the decay of democracy.
\newblock {\em The Virginia Quarterly Review}, 37(4):481--494, 1961.

\bibitem{wlezien2005salience}
Christopher Wlezien.
\newblock On the salience of political issues: The problem with ‘most important problem’.
\newblock {\em Electoral studies}, 24(4):555--579, 2005.

\bibitem{velez_2023_tradeoffs_in_latino_politics}
Yamil~Ricardo Velez.
\newblock Trade-offs in latino politics: Exploring the role of deeply-held issue positions using a dynamic tailored conjoint method.
\newblock {\em Aletheia}, 2023.

\bibitem{stagnaro2024representativeness}
Michael~Nicholas Stagnaro, James Druckman, Adam~J Berinsky, Antonio~Alonso Arechar, Robb Willer, and David Rand.
\newblock Representativeness versus attentiveness: A comparison across nine online survey samples.
\newblock {\em PsyArXiv}, 2024.

\bibitem{agrawal2017near}
Shipra Agrawal and Navin Goyal.
\newblock Near-optimal regret bounds for thompson sampling.
\newblock {\em Journal of the ACM (JACM)}, 64(5):1--24, 2017.

\bibitem{mellon2024ais}
Jonathan Mellon, Jack Bailey, Ralph Scott, James Breckwoldt, Marta Miori, and Phillip Schmedeman.
\newblock Do ais know what the most important issue is? using language models to code open-text social survey responses at scale.
\newblock {\em Research \& Politics}, 11(1):20531680241231468, 2024.

\bibitem{cortina2022conspiratorial}
Jeronimo Cortina and Brandon Rottinghaus.
\newblock Conspiratorial thinking in the latino community on the 2020 election.
\newblock {\em Research \& Politics}, 9(1):20531680221083535, 2022.

\bibitem{velez2023latino}
Yamil~R Velez, Ethan Porter, and Thomas~J Wood.
\newblock Latino-targeted misinformation and the power of factual corrections.
\newblock {\em The Journal of Politics}, 85(2):789--794, 2023.

\bibitem{chen2023prompting}
Boqi Chen, Fandi Yi, and D{\'a}niel Varr{\'o}.
\newblock Prompting or fine-tuning? a comparative study of large language models for taxonomy construction.
\newblock In {\em 2023 ACM/IEEE International Conference on Model Driven Engineering Languages and Systems Companion (MODELS-C)}, pages 588--596. IEEE, 2023.

\bibitem{hopkins2018increasingly}
Daniel~J Hopkins.
\newblock {\em The increasingly United States: How and why American political behavior nationalized}.
\newblock University of Chicago Press, 2018.

\bibitem{weller1998structured}
Susan~C Weller.
\newblock Structured interviewing and questionnaire construction.
\newblock {\em Handbook of methods in cultural anthropology}, pages 365--409, 1998.

\bibitem{egami2018make}
Naoki Egami, Christian~J Fong, Justin Grimmer, Margaret~E Roberts, and Brandon~M Stewart.
\newblock How to make causal inferences using texts.
\newblock {\em arXiv preprint arXiv:1802.02163}, 2018.

\bibitem{offer2022battling}
Molly Offer-Westort, Leah~R Rosenzweig, and Susan Athey.
\newblock Battling the coronavirusinfodemic'among social media users in africa.
\newblock {\em arXiv preprint arXiv:2212.13638}, 2022.

\bibitem{qin2022adaptive}
Chao Qin and Daniel Russo.
\newblock Adaptive experimentation in the presence of exogenous nonstationary variation.
\newblock {\em arXiv preprint arXiv:2202.09036}, 2022.

\bibitem{cavenaghi2021non}
Emanuele Cavenaghi, Gabriele Sottocornola, Fabio Stella, and Markus Zanker.
\newblock Non stationary multi-armed bandit: Empirical evaluation of a new concept drift-aware algorithm.
\newblock {\em Entropy}, 23(3):380, 2021.

\bibitem{lenz2012follow}
Gabriel~S Lenz.
\newblock {\em Follow the leader?: how voters respond to politicians' policies and performance}.
\newblock University of Chicago Press, 2012.

\end{thebibliography}

\end{document}